\documentclass[runningheads]{llncs}
\usepackage{graphicx}
\usepackage{todonotes}
\usepackage{multirow}

\usepackage{array}
\usepackage{tabu}
\newcolumntype{M}[1]{>{\centering\arraybackslash}m{#1}}
\newcolumntype{L}[1]{>{\arraybackslash}m{#1}}

\begin{document}
\title{Overview of BioASQ 2023: The eleventh BioASQ challenge on Large-Scale Biomedical Semantic Indexing and Question Answering
}
    \titlerunning{Overview of BioASQ 2023}
%

\author{
Anastasios Nentidis\inst{1,2} \and
Georgios Katsimpras\inst{1} \and
Anastasia Krithara\inst{1} \and 
Salvador Lima López\inst{3} \and 
Eulália Farré-Maduell\inst{3} \and
Luis Gasco\inst{3} \and
Martin Krallinger\inst{3} \and
Georgios Paliouras\inst{1}
}
\authorrunning{A. Nentidis et al.}
%
\institute{
National Center for Scientific Research ``Demokritos'', Athens, Greece\\
\email{\{tasosnent, gkatsibras, akrithara, paliourg\}@iit.demokritos.gr}\\
\and
Aristotle University of Thessaloniki, Thessaloniki, Greece\\ \and
Barcelona Supercomputing Center, Barcelona, Spain\\
\email{\{salvador.limalopez, eulalia.farre, lgasco, martin.krallinger\}@bsc.es}
}
\maketitle              
\begin{abstract}
This is an overview of the eleventh edition of the BioASQ challenge in the context of the Conference and Labs of the Evaluation Forum (CLEF) 2023. 
BioASQ is a series of international challenges promoting advances in large-scale biomedical semantic indexing and question answering. 
This year, BioASQ consisted of new editions of the two established tasks b and Synergy, and a new task (MedProcNER) on semantic annotation of clinical content in Spanish with medical procedures, which have a critical role in medical practice. 
In this edition of BioASQ, 28 competing teams submitted the results of more than 150 distinct systems in total for the three different shared tasks of the challenge. 
Similarly to previous editions, most of the participating systems achieved competitive performance, suggesting the continuous advancement of the state-of-the-art in the field.  

\end{abstract}
\section{Introduction}
The BioASQ challenge has been focusing on the advancement of the state-of-the-art in large-scale biomedical semantic indexing and question answering (QA) for more than 10 years~\cite{Tsatsaronis2015}. 
In this direction, it organizes different shared tasks annually, developing respective benchmark datasets that represent the real information needs of experts in the biomedical domain. 
This allows the participating teams from around the world, who work on the development of systems for biomedical semantic indexing and question answering, to benefit from the publicly available datasets, evaluation infrastructure, and exchange of ideas in the context of the BioASQ challenge and workshop. 

Here, we present the shared tasks and the datasets of the eleventh BioASQ challenge in 2023, as well as an overview of the participating systems and their performance.
The remainder of this paper is organized as follows. 
First, Section~\ref{sec:tasks} presents a general description of the shared tasks, which took place from January to May 2023, and the corresponding datasets developed for the challenge. 
Then, Section~\ref{sec:participants} provides a brief overview of the participating systems for the different tasks. 
Detailed descriptions for some of the systems are available in the proceedings of the lab. 
Subsequently, in Section~\ref{sec:results}, we present the performance of the systems for each task, based on state-of-the-art evaluation measures or manual assessment.
Finally, in Section~\ref{sec:conclusion} we draw some conclusions regarding the 2023 edition of the BioASQ challenge.

\section{Overview of the tasks}
\label{sec:tasks}
The eleventh edition of the BioASQ challenge (BioASQ 11) consisted of three tasks: (1) a biomedical question answering task (task b), (2) a task on biomedical question answering on developing medical problems (task Synergy), both considering documents in English, and (3) a new task on semantic annotation of medical documents in Spanish with clinical procedures (MedProcNER)~\cite{nentidis2023bioasq}. In this section, we first describe this year's editions of the two established tasks b (task 11b) and Synergy (Synergy 11) with a focus on differences from previous editions of the challenge~\cite{nentidis2022overview,nentidis2021overview}. Additionally, we also present the new MedProcNER task on clinical procedure semantic recognition, linking, and indexing in Spanish medical documents.

\subsection{Biomedical semantic QA - task 11b}
The eleventh edition of task b (task 11b) focuses on a large-scale question-answering scenario in which the participants are required to develop systems for all the stages of biomedical question answering. 
As in previous editions, the task examines four types of questions: “yes/no”, “factoid”, “list” and “summary” questions \cite{balikas13}.
In this edition, the training dataset provided to the participating teams for the development of their systems consisted of 4,719 biomedical questions from previous versions of the challenge annotated with ground-truth relevant material, that is, articles, snippets, and answers~\cite{krithara2023bioasq}. 
Table \ref{tab:b_data} shows the details of both training and test datasets for task 11b.
The test data for task 11b were split into four independent bi-weekly batches. These include two batches of 75 questions and two batches of 90 questions each, as presented in Table \ref{tab:b_data}. 

\begin{table}[!htb]
        \caption{Statistics on the training and test datasets of task 11b. The numbers for the documents and snippets refer to averages per question.}\label{tab:b_data}
        \centering
        \begin{tabular}{M{0.09\linewidth}M{0.08\linewidth}M{0.08\linewidth}M{0.09\linewidth}M{0.15\linewidth}M{0.15\linewidth}M{0.15\linewidth}M{0.15\linewidth}}\hline
        \textbf{Batch} 	& \textbf{Size} 	&	\textbf{Yes/No}	&\textbf{List}	&\textbf{Factoid}	&\textbf{Summary}& \textbf{Documents} 	& \textbf{Snippets}  	\\ \hline
        Train   & 4,719 & 1,271 & 901 & 1,417 & 1,130 & 9.03 & 12.04 \\
        Test 1	& 75   & 24   & 12  & 19   & 20   & 2.48 & 3.28  \\
        Test 2	& 75   & 24   & 12  & 22   & 17   & 2.96 & 4.33  \\
        Test 3	& 90   & 24   & 18  & 26   & 22   & 2.66 & 3.77  \\
        Test 4	& 90   & 14   & 24  & 31   & 21   & 2.80 & 3.91  \\\hline  
        \textbf{Total} & 5,049 & 1,357 & 967 & 1,515 & 1,210 & 8.62 & 11.5  \\\hline  
        \end{tabular}
\end{table}

As in previous editions of task b, task 11b was also divided into two phases which run for two consecutive days for each batch: (phase A) the retrieval of the relevant material and (phase B) providing the answers to the questions.
In each phase, the participants have 24 hours to submit the responses generated by their systems.
In particular, a test set consisting of the bodies of biomedical questions, written in English, was released for phase A and the participants were expected to identify and submit relevant elements from designated resources, namely PubMed/MEDLINE-article abstracts, and snippets extracted from these resources.
Then, some relevant articles and snippets for these questions, which have been manually selected by the experts, were also released in phase B and the participating systems were challenged to respond with \textit{exact answers}, that is entity names or short phrases, and \textit{ideal answers}, that is, natural language summaries of the requested information.

\subsection{Task Synergy 11}
\begin{figure*}[!htb]
\centerline{\includegraphics[width=0.7\textwidth]{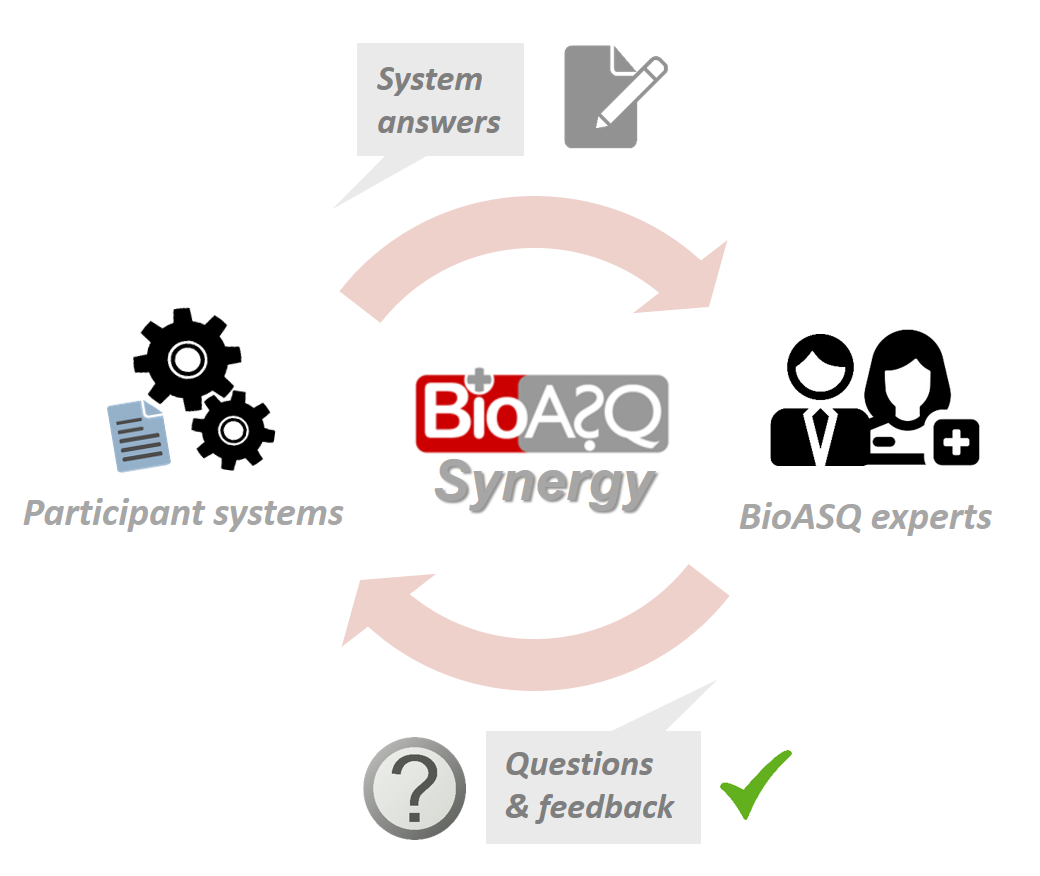}}
\caption{The iterative dialogue between the experts and the systems in the BioASQ Synergy task on question answering for developing biomedical problems. }\label{fig:synergy}
\end{figure*}

The task Synergy was introduced two years ago~\cite{nentidis2021overview} envisioning a continuous dialog between the experts and the systems.
In task Synergy, the motivation is to make the advancements of biomedical information retrieval and question answering available to biomedical experts studying open questions for developing problems, aiming at a synergy between automated question-answering systems and biomedical experts. 
In this model, the systems provide relevant material and answers to the experts that posed some open questions. The experts assess these responses and feed their assessment back to the systems. 
This feedback is then exploited by the systems in order to provide more relevant material, considering more recent material that becomes available in the meantime, and improved responses to the experts as shown in Figure {\ref{fig:synergy}}.  
This process proceeds with new feedback and new responses from the systems for the same open questions that persist, in an iterative way, organized in rounds. 

 After eight rounds of the task Synergy in the context of BioASQ9~\cite{krithara2021bioasq} and four more in the context of BioASQ10~\cite{nentidis2022bioasq}, all focusing on open questions for the developing problem of the COVID-19 pandemic, in BioASQ11 we extended the Synergy task (Synergy 11) to open questions for any developing problem of interest for the participating biomedical experts~\cite{nentidis2023bioasq}. 
 In this direction, the four bi-weekly rounds of Synergy 11 were open to any developing problem, and a designated version of the PubMed/MEDLINE repository was considered for the retrieval of relevant material in each round. As in previous versions of the task, and contrary to task b, the open questions were not required to have definite answers and the answers to the questions could be more volatile.
 In addition, a set of 311 questions on COVID-19, from the previous versions of the Synergy task, were available, together with respective incremental expert feedback and answers, as a development set for systems participating in this edition of the task.   
 Table \ref{tab:syn_data} shows the details of the datasets used in task Synergy 11.
 
\begin{table}[!htb]
	\caption{Statistics on the datasets of Task Synergy. ``Answer ready'' stands for questions marked as having enough relevant material to be answered after the assessment of material submitted by the systems in the respective round. In round 2, ten questions were omitted from the test, as no feedback was available for them from the respective expert for the material retrieved by the systems in round 1. This feedback become available in round three, hence these questions were again included in the test set. }\label{tab:syn_data}
        \centering
        \begin{tabular}{c c c c c c c c}\hline
      	 \textbf{Round}  & \textbf{Size}  & \textbf{Yes/No} &\textbf{List} &\textbf{Factoid} &\textbf{Summary}& \textbf{Answer  ready}  \\
		\hline
 1  & 53 & 12 & 17 & 11 & 13 & 14 \\
 2  & 43 & 11 & 14 & 7 & 11 & 32 \\
 3  & 53 & 12 & 17 & 11 & 13 & 37 \\
 4  & 53 & 12 & 17 & 11 & 13 & 42 \\
	 \hline                 
	\end{tabular}
\end{table}
 
 Similar to task 11b, four types of questions are examined in Synergy 11 task: yes/no, factoid, list, and summary, and two types of answers, \textit{exact} and \textit{ideal}. Moreover, the assessment of the systems' performance is based on the evaluation measures used in task 11b.
 However, contrary to task 11b, Synergy 11 was not structured into phases, with both relevant material and answers received together. 
 For new questions, only relevant material, that is relevant articles and snippets, was required until the expert considered that enough material has been gathered and marked the questions as ``ready to answer". 
 Once a question is marked as ``ready to answer", the systems are expected to respond to the experts with both new relevant material and answers in subsequent rounds.

\subsection{Medical semantic annotation in Spanish - MedProcNER}

Clinical procedures play a critical role in medical practice, being an essential tool for the diagnosis and treatment of patients. They are also a difficult information type to extract, often being made up of abbreviations, multiple parts, and even descriptive sections. Despite their importance, there are not many resources that focus in-depth on the automatic detection of clinical procedures, and even fewer, if any, consider concept normalization.

\begin{figure*}[!htb]
\centerline{\includegraphics[width=1\textwidth]{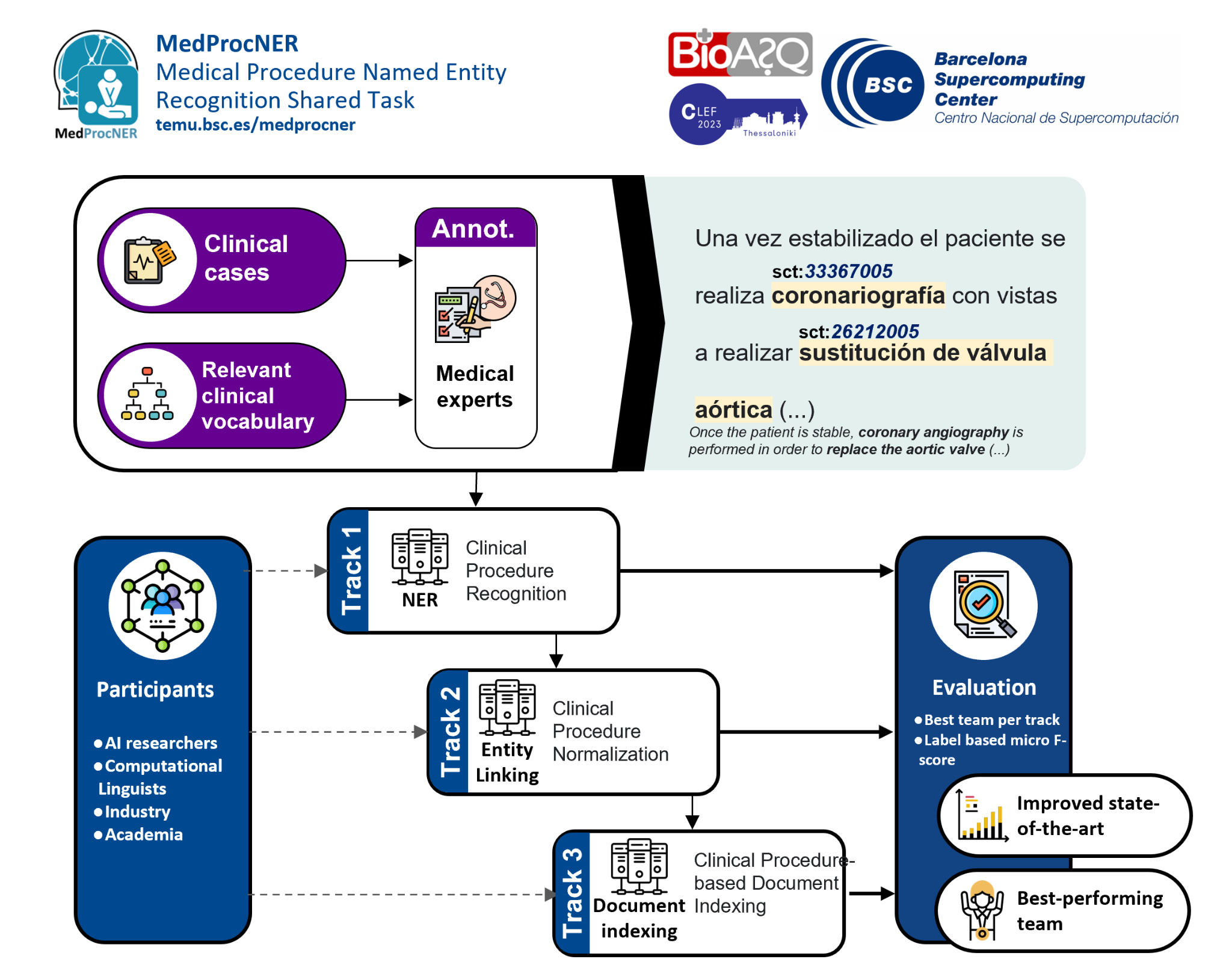}}
\caption{Overview of the MedProcNER Shared Task.}\label{fig:medprocneroverview}
\end{figure*}

With this in mind, this year we introduced the MedProcNER (Medical Procedure Named Entity Recognition) shared task as part of BioASQ11 as summarized in Figure~\ref{fig:medprocneroverview}. The task challenges participants to create automatic systems that can extract different aspects of information about clinical procedures. These aspects are divided into three different sub-tasks: 

\begin{itemize}
    \item \textbf{Clinical Procedure Recognition:} This is a named entity recognition (NER) task where participants are challenged to automatically detect mentions of clinical procedures in a corpus of clinical case reports in Spanish.
    \item \textbf{Clinical Procedure Normalization:} In this entity linking (EL) task, participants must create systems that are able to assign SNOMED CT codes to the mentions retrieved in the previous sub-task.
    \item \textbf{Clinical Procedure-based Document Indexing:} This is a semantic indexing challenge in which participants automatically assign clinical procedure SNOMED CT codes to the full clinical case report texts so that they can be indexed. In contrast to the previous sub-task, participants do not need to rely on any previous systems, making this an independent sub-task.
\end{itemize}

To enable the development of clinical procedure recognition, linking and indexing systems, we have released the MedProcNER/ProcTEMIST corpus, a Gold Standard dataset of 1,000 clinical case reports manually annotated by multiple clinical experts with clinical procedures. The case reports were carefully selected by clinical experts and belong to various medical specialties including, amongst others, oncology, odontology, urology, and psychiatry. They are the same text documents that were used for the corpus and shared task on diseases DisTEMIST \cite{amiranda2022overview}, building towards a collection of fully-annotated texts for clinical concept recognition and normalization. 
The MedProcNER corpus is publicly available on Zenodo\footnote{https://doi.org/10.5281/zenodo.7817745}.

In addition to the text annotations, the mentions in the corpus have been normalized to SNOMED CT. SNOMED CT (Systematized Nomenclature of Medicine Clinical Terms) is a comprehensive clinical terminology and coding system designed to facilitate the exchange and communication of health-related information across different healthcare settings and systems, which makes it fit for the normalization of varied clinical concepts. For the task, only a subset of 250 normalized documents was released as training data. The complete normalized dataset will be released as post-workshop material.

Annotation and normalization guidelines were specifically created for this task. 
The current version of the guidelines includes 31 pages and a total of 60 rules that describe how to annotate different procedure types ranging from simple explorations to complex surgical descriptions. They also include a discussion of the task’s importance and use cases, basic information about annotation process, a description of different procedure types and comparisons with similar clinical entity types, and indications and resources for the annotators.
As with the DisTEMIST corpus, the guidelines were refined via multiple rounds of inter-annotator agreement (IAA) through parallel annotation of a section of the corpus. The final IAA score (computed as the pairwise agreement between two independent annotators) is of 81.2. The MedProcNER guidelines are available in Zenodo\footnote{https://doi.org/10.5281/zenodo.7817666}.

In addition to the corpus and guidelines, some additional resources have been released as part of the task. First, a SNOMED CT gazetteer was released containing official terms and synonyms from the relevant branches of SNOMED CT for the grounding of procedure mentions. The MedProcNER gazetteer has been built using the 31/10/2022 version of the Spanish edition of SNOMED CT, which is composed than 300,000 concepts organized in 19 different hierarchies including ``procedure", ``substance" and "regime/therapy". To simplify the entity linking and indexing task, we compiled a reduced subset of the terminology with a smaller set of concepts to which the mentions can be mapped. The gazetteer consists of 234,674 lexical entries, out of which 130,219 are considered main terms. Within these entries, there are 130,219 unique codes originating from 19 hierarchies. 

Next, to foster the advancement of document indexing with other terminologies and boost the reusability of MedProcNER data, we have created cross-mappings that connect the SNOMED CT mentions found in the corpus to MeSH and ICD-10. These mappings were achieved using the UMLS Meta-thesaurus. 
Finally, a Multilingual Silver Standard similar to last year's DisTEMIST \cite{amiranda2022overview} and LivingNER \cite{livingner} was created in six different languages: English, French, Italian, Portuguese, Romanian and Catalan. This Silver Standard was automatically generated using a lexical annotation transfer approach in which the corpus' texts and Gold Standard annotations are translated separately and then mapped onto each other using a look-up system. This look-up takes into account individual annotations in each file, their translations and also a lemmatized version of the entities (obtained using spaCy\footnote{https://spacy.io/}). Transferred annotations carry over the SNOMED CT code originally assigned to the Spanish annotation. All additional resources are available in Zenodo together with the Gold Standard corpus\footnote{https://doi.org/10.5281/zenodo.7817666}.

As for the task evaluation, all three MedProcNER sub-tasks are evaluated using micro-averaged precision, recall and F1-score. It is important to highlight that the evaluation of entity linking systems is not conducted in isolation but rather in an end-to-end manner. Instead of being provided an exhaustive list of mentions to be normalized, participants had to rely on their predictions from the named entity recognition stage. Consequently, the obtained scores might not accurately represent the overall performance of the systems. However, this type of evaluation does offer a more comprehensive assessment of complete systems, closely resembling their performance in real-world applications.
 
MedProcNER is promoted by the Spanish Plan for the Advancement of Language Technology (Plan TL)\footnote{https://plantl.mineco.gob.es} and organized by the Barcelona Supercomputing Center (BSC) in collaboration with BioASQ. A more in-depth analysis of the MedProcNER Gold Standard, guidelines and additional resources is presented in the MedProcNER overview paper \cite{medprocner}.

\section{Overview of participation}
\label{sec:participants}
\subsection{Task 11b}

19 teams competed this year in task 11b submitting the responses of 76 different systems for both phases A and B, in total. In particular, 9 teams with 37 systems participated in Phase A, while in Phase B, the number of participants and systems were 16 and 59 respectively. Six teams engaged in both phases.
An overview of the technologies employed by the teams is provided in Table \ref{tab:b_sys} for the systems for which a description was available. Detailed descriptions for some of the systems are available at the proceedings of the workshop.

\begin{table}[!htb]
        \centering
         \caption{Systems and approaches for task 11b. Systems for which no information was available at the time of writing are omitted.}
        \begin{tabular}{M{0.23\linewidth}M{0.1\linewidth}M{0.05\linewidth}M{0.56\linewidth}}\hline
        \textbf{Systems} & \textbf{Phase}& \textbf{Ref}& \textbf{Approach} \\ \hline
        bioinfo	 & A, B & \cite{Almeida2023} & BM25, PubMedBERT, monoT5, reciprocal rank fusion (RRF), ALPACA-LoRA, OA-Pythia, OA-LLaMA  \\\hline 
        UR-gpt & A, B & \cite{Ateia2023} & GPT-3.5-turbo, GPT-4\\\hline 
        ELECTROBERT	 & A, B & \cite{Panou2023} & ELECTRA, ALBERT, BioELECTRA, BERT, GANBERT, BM25, RM3  \\\hline 
        MindLab & A, B & \cite{RossoMateus2023} & BM25, CNN, SBERT \\\hline 
        dmiip & A, B & - & BM25, GPT-3.5, PubMedBERT, BioBERT, BioLinkBERT, ELECTRA \\\hline 

        
        A\&Q & A  & \cite{Shin2023}& BM25, PubMedBERT \\\hline 
        IRCCS & A  & - & transformers, cosine similarity, BM25 \\\hline 
        MarkedCEDR & A & \cite{Lesavourey2023} &  BM25, BERT, CEDR \\\hline 


        ELErank & B & \cite{Aksenova2023}& BioM-ELECTRA, S-PubMedBERT\\\hline  
        NCU-IISR & B & \cite{Hsueh2023} & GPT-3, GPT-4 \\\hline 
        AsqAway & B & \cite{RaghavR2023} & BioBERT, BioM-Electra \\\hline  
        MQ & B & - & GPT-3.5, sBERT, DistilBERT \\\hline 
        DMIS-KU & B & \cite{HyunjaeKim2023} & BioLinkBERT, GPT-4, data augmentation \\\hline 
        MQU & B & \cite{Galat2023} & BART, BioBART \\\hline 

        
        \end{tabular}
       \label{tab:b_sys}
\end{table}

The (``\textit{bioinfo}'') team from the University of Aveiro participated in both phases of the task with five systems. In phase A, they developed a two-stage retrieval pipeline. The first stage adopted the traditional BM25 model. In contrast, the second stage implemented transformer-based neural re-ranking models from PubMedBERT and monoT5 checkpoints. Additionally, synthetic data were used to augment the training regimen. The reciprocal rank fusion (RRF) was utilized to ensemble the outputs from various models.
For Phase B, their systems utilized instruction-based transformer models, such as ALPACA-LoRA, OA-Pythia, and OA-LLaMA, for conditioned zero-shot answer generation. More specifically, given the most relevant article from Phase A, the model was designed to generate an \textit{ideal answer} based on the information contained in the relevant article.

Another team participating in both phases is the team from the University of Regensburg. Their systems (``\textit{UR-gpt}'') relied on two commercial versions of the GPT Large Language model (LLM). Specifically, their systems experimented with both GPT-3.5-turbo and GPT-4 models. In phase A, their systems used zero-shot learning for query expansion, query reformulation and re-ranking. For Phase B, they used zero-shot learning, grounded with relevant snippets.

The BSRC Alexander Fleming team also participated in both phases with the systems``\textit{ELECTROBERT}''. Their systems are built upon their previously developed systems~\cite{reczko2022electrolbert} and also adapted the semi-supervised method GANBERT~\cite{croce2020gan} for document relevance classification. Furthermore, for the initial document selection phase their systems utilize BM25 combined with RM3 query expansion with optimized parameters.  

The `\textit{MindLab}'' team competed in both phases of the task with five systems. For document retrieval their systems used the BM25 scoring function and semantic-similarity as a re-ranking strategy. For passage retrieval their systems used a metric learning method which fuses different
similarity measures through a siamese convolutional network.

The ``\textit{dmiip}'' team from the Fudan University participated in both phases of the task with five systems. In phase A, their systems used BM25 and GPT for the retrieval stage, and a cross-encoder ranker based on different biomedical PLMs, such as PubMedBERT, BioBERT, BioLinkBERT and ELECTRA for the ranking stage.  Biomedical PLMs and GPT-3.5 are also utilized in Phase B. The systems are initially finetuned on SQuAD and then trained with the BioASQ traning dataset.

In phase A, the ``\textit{A\&Q}'' team participated with five systems. Their systems are based on a multi-stage approach which incorporates a bi-encoder model in the retrieval stage, and
a cross-encoder model at the re-ranking stage. At the retrieval stage, a hybrid retriever that combines dense and sparse retrieval, where the dense retrieval is implemented with the bi-encoder and the sparse retrieval is implemented with BM25. Both encoders are initialized with PubMedBERT and further trained on PubMed query-article search logs.

The IRCCS team participated with five systems (``\textit{IRCCS}'') in phase A. Their systems follow a two-step methodology. First, they score the documents using the BM25 ranking function. Then, the second step is to re-rank them  based on cosine similarity between the query and each document, which are encoded using various transformers models.

The IRIT lab team competed also in phase A with two systems (``\textit{MarkedCEDR}''). Their systems adopt a two-stage retrieval approach composed of a retriever and a re-ranker. The former is based on BM25. The later is an implementation of a BERT cross-encoder named CEDR.

In phase B, the Ontotext team participated with two systems (``\textit{ELErank}'').
Their systems used BioM-ELECTRA as a backbone model for both yes/no and factoid questions. For yes/no questions, it was fine-tuned in a sequence classification setting, and for factoid questions, it was fine-tuned in a token classification setting (for extractive QA). Before applying classification, the sentences were ranked based on their cosine similarity to the question. Top-5 most relevant sentences were used for classification. Sentence embeddings for ranking were calculated with S-PubMedBERT.

The National Central Uni team competed with five systems ``\textit{NCU-IISR}'' in phase B. Their systems utilized OpenAI’s ChatCompletions API, incorporating Prompt Engineering techniques to explore various prompts. Specifically, their systems used GPT-3 and GPT-4 for answer generation.
    
The CMU team participated with four different systems (``\textit{AsqAway}'') in phase B. Their system adopt an ensembling approach using transformer models. For factoid and list questions they use a BioBERT and BioM-Electra ensemble. For yes/no questions, they employ BioM-Electra. 

The Korea University team participated with five systems (``\textit{DMIS-KU}''). They employed different pre-processing, training, and data augmentation methods and different QA models. For the yes/no type, the systems utilized the ``full-snippet" pre-processing method, where all snippets were concatenated into a single context. The BioLinkBERT-large model was used as the embedding model. For the factoid type, the ``single-snippet" method was used, which involved processing one snippet at a time. The BioLinkBERT-large was trained using the SQuAD dataset and fine-tuned using the BioASQ training data. For the list type, the full-snippet method was used again. Additionally, their systems employed a dataset generation framework, called LIQUID, to augment the training data. Also, the GPT-4 was utilized to answer list questions in a one-shot manner. In all question types, the final predictions are produced by combining the results from multiple single models using an ensemble method.

There were two teams from the  Macquarie University. The first team participated with five systems (``\textit{MQ}'')  in phase B and focused on finding the \textit{ideal answers}. Three of their systems employed GPT-3.5 and various types of prompts. The rest of the systems were based on their previously developed systems~\cite{molla2022query}.
The second tean (``\textit{MQU}'') competed with five systems in phase B which utilised BART and BioBART that were fine-tuned for abstractive summarisation.


As in previous editions of the challenge, a baseline was provided for phase B \textit{exact answers}, based on the open source OAQA system\cite{yang2016learning}. This system relies on more traditional NLP and Machine Learning approaches, used to achieve top performance in older editions of the challenge, and now serves as a baseline. The system is developed based on the UIMA framework. In particular, question and snippet parsing is done with ClearNLP. Then, MetaMap, TmTool \cite{Wei2016}, C-Value, and LingPipe \cite{baldwin2003lingpipe} are employed for identifying concepts that are retrieved from the UMLS Terminology Services (UTS). Finally, the relevance of concepts, documents, and snippets is identified based on some classifier components and some scoring and ranking techniques are also employed.

Furthermore, this year we introduced two more baselines for phase B \textit{ideal answers}, BioASQ Baseline ZS and BioASQ Baseline FS, which are based on zero-shot prompting of Biomedical LMs. Both systems utilized the BioGPT, a language model trained exclusively on biomedical abstracts and papers, with the former using as input only the question body, and the latter using the concatenation of the question body and the relevant snippets until the input length is exceeded.

\subsection{Task Synergy 11}

In this edition of the task Synergy (Synergy 11) 5 teams participated submitting the results from 12 distinct systems. 
An overview of systems and approaches employed in this task is provided in Table \ref{tab:syn}, for the systems for which a description was available. More detailed descriptions for some of the systems are available at the proceedings of the workshop.

\begin{table}[!htb]
        \centering
         \caption{Systems and their approaches for task Synergy. Systems for which no description was available at the time of writing are omitted. }
        \begin{tabular}{M{0.2\linewidth}M{0.06\linewidth}M{0.7\linewidth}}\hline
        \textbf{System} & \textbf{Ref} & \textbf{Approach} \\ \hline
        dmiip  & - & BM25, GPT-3.5, PubMedBERT, BioBERT, BioLinkBERT, ELECTRA\\\hline 
        bio-answerfinder & \cite{ozyurt2021end} & Bio-ELECTRA++, BERT, weighted relaxed word mover's distance (wRWMD), pyserini with MonoT5, SQuAD, GloVe \\\hline 
        ELECTROBERT	 & \cite{Panou2023} & ELECTRA, ALBERT, BioELECTRA, BERT, GANBERT, BM25, RM3  \\\hline 
        \end{tabular}
        \label{tab:syn}
\end{table}

The Fudan University (``\textit{dmipp}'') competed in task Synergy with the same models they used for task 11b. Additionally, they expanded the query with the shortest relevant snippet in the provided feedback.

The ``\textit{UCSD}'' team competed in task Synergy with two systems. Their systems (``\textit{bio-answerfinder}'') used the Bio-AnswerFinder end-to-end QA system they had previously developed \cite{ozyurt2021end} with few improvements, including the use of the expert feedback data in retraining of their model's re-ranker.

The BSRC Alexander Fleming team participated with two systems. Similar to task b, their systems (``\textit{ELECTROBERT}'') built upon their previously developed systems~\cite{reczko2022electrolbert} and also adapted the semi-supervised method GANBERT~\cite{croce2020gan}.

\subsection{Task MedProcNER}
Among the 47 teams registered for the MedProcNER task, 9 teams submitted at least one run of their predictions. Specifically, all 9 teams engaged in the entity recognition sub-task, while 7 teams participated in the entity linking sub-task. Additionally, 4 teams took part in the document indexing sub-task. Overall, a total of 68 runs were submitted, reflecting the collective efforts and contributions of the participating teams.

\begin{table}[!htb]
        \centering
         \caption{Systems and approaches for task MedProcNER. Systems for which no description was available at the time of writing are omitted. In the Task column, NER stands for named entity recognition (i.e. sub-task 1), EL for entity linking (i.e. sub-task 2) and DI for document indexing (i.e. sub-task 3)}
        \begin{tabular}{M{0.24\linewidth}M{0.05\linewidth}M{0.07\linewidth}M{0.62\linewidth}}\hline
        \textbf{Team} & \textbf{Ref} & \textbf{Task} & \textbf{Approach} \\ \hline
         BIT.UA & \cite{bituaMedprocner} & NER  & Transformer-based solution with masked CRF and data augmentation \\
         & & EL & Semantic search with pretrained transformer-based models using an unsupervised approach \\ 
         & & DI & Indexing of codes found in EL step \\
         \midrule
         Fusion & \cite{fusionmedprocner} & NER & Different BERT-family models fine-tuned for token classification \\
         & & EL & Cross-lingual SapBERT (SapBERT-UMLS-2020AB-all-lang-from-XLMR-large) \\
         \midrule
         KFU NLP Team & - & NER & Ensemble of different SapBERT and mGEBERT models with and without adapters \\
         & & EL & Synonym Marginalization loss function and UniPELT adapters \\
         & & DI & Indexing of codes found in EL step \\
         \midrule
         NLP-CIC-WFU & - & NER & Fine-tuned RoBERTa models combined with different pre-processing and post-processing techniques \\
         \midrule
         Onto-NLP & \cite{ontomedprocner} & NER &  Fine-tuned RoBERTa models + lexical matching \\
         & & EL & SapBERT models + lexical matching + majority voting \\
         \midrule
         University of Regensburg & \cite{Ateia2023} & All & In-context few-shot (3) learning with GPT-3.5-turbo and GPT-4 \\
         \midrule
         SINAI & \cite{sinaimedprocner} & NER & Clinical transformer models + recurrent classifiers (GRU, CRF) \\
         & & EL & Combination of matching techniques with token similarity based normalization \\
         \midrule
         Vicomtech & \cite{vicommedprocner} & NER & Seq2seq systems with BERT-like models \\
         & & EL & Semantic search techniques based on transformer models (SapBERT) and cross-encoders \\
         & & DI & Combination of first two systems \\
         \bottomrule
         \end{tabular}
        \label{tab:medprocner_sys}
\end{table}

Table \ref{tab:medprocner_sys} gives an overview of the methodologies used by the participants in each of the sub-tasks. 
As is the case in many modern NLP approaches, the majority of the participants used transformers-based models. RoBERTa \cite{roberta} and SapBERT \cite{sapbert} models were the most popular for named entity recognition and entity linking respectively. In addition to this, in order to boost the systems' performance some teams also relied on recurrent classifiers such as CRFs (e.g. BIT.UA \cite{bituaMedprocner}, SINAI team \cite{sinaimedprocner}), adapters (e.g. KFU NLP team), model ensembling/voting (e.g. KFU NLP team, Onto-text \cite{ontomedprocner}) and data augmentation (e.g. BIT.UA \cite{bituaMedprocner}). Interestingly, one of the participants (Samy Ateia from the University of Regensburg \cite{Ateia2023}) proposes an approach based on Generative Pre-trained Transformers (GPT) models for all three sub-tasks.

\section{Results}
\label{sec:results}

\subsection{Task 11b}
\textbf{Phase A}: 
The Mean Average Precision (MAP) was the official measure for evaluating system performance on document retrieval in phase A of task 11b, which is based on the number of ground-truth relevant elements.
For snippet retrieval, however, the situation is more complicated as a ground-truth snippet may overlap with several distinct submitted snippets, which makes the interpretation of MAP less straightforward. 
For this reason, since BioASQ9 the F-measure is used for the official ranking of the systems in snippet retrieval, which is calculated based on character overlaps\footnote{\url{http://participants-area.bioasq.org/Tasks/b/eval\_meas\_2022/}}~\cite{nentidis2021overview}.

Since BioASQ8, a modified version of Average Precision (AP) is adopted for MAP calculation. 
In brief, since BioASQ3, the participant systems are allowed to return up to 10 relevant items (e.g. documents or snippets), and the calculation of AP was modified to reflect this change. However, some questions with fewer than 10 golden relevant items have been observed in the last years, resulting in relatively small AP values even for submissions with all the golden elements. Therefore, the AP calculation was modified to consider both the limit of 10 elements and the actual number of golden elements~\cite{nentidis2020overview}.  

\begin{table*}[!htbp]
\centering
\caption{Preliminary results for document retrieval in batch 1 of phase A of task 11b. 
Only the top-10 systems are presented, based on MAP.
}
\begin{tabular}{M{0.3\linewidth}M{0.13\linewidth}M{0.13\linewidth}M{0.13\linewidth}M{0.13\linewidth}M{0.12\linewidth}}\hline
\textbf{System} & \textbf{Mean Precision} & \textbf{Mean Recall} & \textbf{Mean F-measure} & \textbf{MAP} & \textbf{GMAP}  \\ \hline
bioinfo-0            & 0.2118         & 0.6047 & \textbf{0.2774} & \textbf{0.4590} & \textbf{0.0267} \\
bioinfo-1           & \textbf{0.2152} & 0.5964    & 0.2769          & 0.4531 & \textbf{0.0267} \\
bioinfo-2            & 0.1498         & 0.5978    & 0.2192          & 0.4522 & 0.0233 \\
bioinfo-3            & 0.1712   & \textbf{0.6149} & 0.2418          & 0.4499 & 0.0239 \\
dmiip3               & 0.1133         & 0.6127    & 0.1823          & 0.4462 & 0.0240 \\
A\&Q4                & 0.1027         & 0.5816    & 0.1667          & 0.4404 & 0.0215 \\
A\&Q3                & 0.1027         & 0.5816    & 0.1667          & 0.4404 & 0.0215 \\
A\&Q5                & 0.1000         & 0.5738    & 0.1627          & 0.4397 & 0.0191 \\
dmiip5               & 0.1120         & 0.5993    & 0.1799          & 0.4391 & 0.0209 \\
bioinfo-4            & 0.1529         & 0.5944    & 0.2183          & 0.4275 & 0.0164 \\
\hline
\\
\end{tabular}
\label{tab:bA_res_doc}

\centering
\caption{Preliminary results for snippet retrieval in batch 1 of phase A of task 11b. Only the top-10 systems are presented, based on F-measure.
        }
\begin{tabular}{M{0.3\linewidth}M{0.14\linewidth}M{0.12\linewidth}M{0.14\linewidth}M{0.12\linewidth}M{0.12\linewidth}}\hline
\textbf{System} & \textbf{Mean Precision} & \textbf{Mean Recall} & \textbf{Mean F-measure} & \textbf{MAP} & \textbf{GMAP}  \\ \hline
dmiip3               & \textbf{0.1109}  & \textbf{0.4309}  & \textbf{0.1647}  & \textbf{0.4535}  & \textbf{0.0104} \\
dmiip4               & 0.1099  & 0.4144  & 0.1628  & 0.4142  & 0.0065 \\
dmiip2               & 0.1075  & 0.4023  & 0.1589  & 0.4234  & 0.0077 \\
dmiip5               & 0.1075  & 0.4053  & 0.1589  & 0.4327  & 0.0089 \\
dmiip1               & 0.1027  & 0.3863  & 0.1518  & 0.4038  & 0.0061 \\
MindLab QA Reloaded  & 0.0833  & 0.1918  & 0.0991  & 0.1389  & 0.0023 \\
MindLab Red Lions++  & 0.0816  & 0.1807  & 0.0944  & 0.1228  & 0.0013 \\
Deep ML methods for  & 0.0808  & 0.1485  & 0.0904  & 0.1208  & 0.0007 \\
MindLab QA System    & 0.0838  & 0.1245  & 0.0887  & 0.0995  & 0.0003 \\
MindLab QA System ++ & 0.0838  & 0.1245  & 0.0887  & 0.0995  & 0.0003 \\
        \hline
        \\
        \end{tabular}
        \label{tab:bA_res_sni}

\centering
\caption{Results for batch 2 for \textit{exact answers} in phase B of task 11b.
Only the top-15 systems based on Yes/No F1 and the BioASQ Baseline are presented.}
\begin{tabular}
{M{0.205\linewidth}M{0.0852\linewidth}M{0.0852\linewidth}M{0.105\linewidth}M{0.11\linewidth}M{0.0852\linewidth}M{0.0852\linewidth}M{0.0852\linewidth}M{0.0852\linewidth}}
\hline

\textbf{System} & \multicolumn{2}{c}{\textbf{Yes/No}} & \multicolumn{3}{c}{\textbf{Factoid}} & \multicolumn{2}{c}{\textbf{List}} \\ 
\hline
& F1 & Acc. & Str. Acc. & Len. Acc. & MRR & Prec. & Rec. & F1 \\ \cline{2-9}    

IISR-2               & \textbf{1.0000} & \textbf{1.0000} & \textbf{0.5455} & 0.6364 & \textbf{0.5909}& \textbf{0.5099} & 0.3577 & 0.3980 \\
IISR-1               & \textbf{1.0000} & \textbf{1.0000} & 0.5000 & 0.5455 & 0.5227 & 0.4861 & 0.3310 & 0.3678 \\
DMIS-KU-4            & \textbf{1.0000} & \textbf{1.0000} & 0.3636 & 0.5909 & 0.4697 & 0.2983 & 0.3683 & 0.2871 \\
DMIS-KU-1            & 0.9524 & 0.9577 & 0.3182 & \textbf{0.6818} & 0.4773 & 0.3349 & 0.3623 & 0.3080 \\
DMIS-KU-2            & 0.9524 & 0.9577 & 0.3182 & \textbf{0.681}8 & 0.4561 & 0.3486 & 0.3456 & 0.3087 \\
DMIS-KU-3            & 0.9524 & 0.9577 & 0.3636 & 0.5909 & 0.4621 & 0.2818 & 0.4058 & 0.3178 \\
UR-gpt4-zero-ret     & 0.9474 & 0.9564 & \textbf{0.5455} & 0.5909 & 0.5682 & 0.3742 & 0.4369 & 0.3828 \\
dmiip5               & 0.9474 & 0.9564 & 0.4091 & 0.4545 & 0.4242 & 0.1413 & 0.2200 & 0.1676 \\
capstone-1           & 0.9091 & 0.9161 & 0.4091 & 0.5455 & 0.4561 & 0.2085 & 0.3810 & 0.2617 \\
DMIS-KU-5            & 0.9000 & 0.9143 & 0.3636 & 0.5909 & 0.4621 & 0.2534 & 0.4593 & 0.3022 \\
dmiip1               & 0.9000 & 0.9143 & 0.3182 & 0.5909 & 0.4318 & 0.2271 & 0.3760 & 0.2501 \\
UR-gpt3.5-turb... & 0.8889 & 0.9111 & \textbf{0.5455} & 0.5909 & 0.5682 & 0.4598 & \textbf{0.4671} & \textbf{0.4316} \\
dmiip3               & 0.8571 & 0.8730 & 0.3182 & 0.5455 & 0.3992 & 0.2851 & 0.2464 & 0.2232 \\
AsqAway\_1           & 0.8421 & 0.8693 & 0.4545 & 0.4545 & 0.4545 & 0.1780 & 0.1968 & 0.1756 \\
AsqAway\_2           & 0.8421 & 0.8693 & 0.4545 & 0.4545 & 0.4545 & 0.2023 & 0.3226 & 0.2327 \\
BioASQ\_Baseline     & 0.6000 & 0.4667 & 0.0909 & 0.1364 & 0.1136 & 0.1185 & 0.2784 & 0.1613 \\
\hline
\end{tabular}

\label{tab:bB_res}
\end{table*}

Tables \ref{tab:bA_res_doc} and \ref{tab:bA_res_sni} present some indicative preliminary results for the retrieval of documents and snippets in batch 1. The full results are available online on the result page of task 11b, phase A\footnote{\footnotesize \url{http://participants-area.bioasq.org/results/11b/phaseA/}}. The final results for task 11b will be available after the  completion of the manual assessment of the system responses by the BioASQ team of biomedical experts, which is still in progress, therefore the results reported here are currently preliminary. 

\textbf{Phase B}: 
In phase B of task 11b, the competing systems submit exact and \textit{ideal answers}. 
As regards the \textit{ideal answers}, the official ranking of participating systems is based on manual scores assigned by the BioASQ team of experts that assesses each \textit{ideal answer} in the responses~\cite{balikas13}.  
The final position of systems providing \textit{exact answers} is based on their average ranking in the three question types where \textit{exact answers} are required, that is ``yes/no'', ``list'', and ``factoid''. Summary questions for which no \textit{exact answers} are submitted are not considered in this ranking.
In particular, the mean F1 measure is used for the ranking in list questions, the mean reciprocal rank (MRR) is used for the ranking in factoid questions, and the F1 measure, macro-averaged over the classes of yes and no, is used for yes/no questions.
Table~\ref{tab:bB_res} presents some indicative preliminary results on \textit{exact answer} extraction from batch 2. The full results of phase B of task 11b are available online\footnote{\footnotesize \url{http://participants-area.bioasq.org/results/11b/phaseB/}}. These results are preliminary, as the final results for task 11b will be available after the manual assessment of the system responses by the BioASQ team of biomedical experts.

\begin{figure*}[!htbp]
\centerline{\includegraphics[width=1\textwidth]{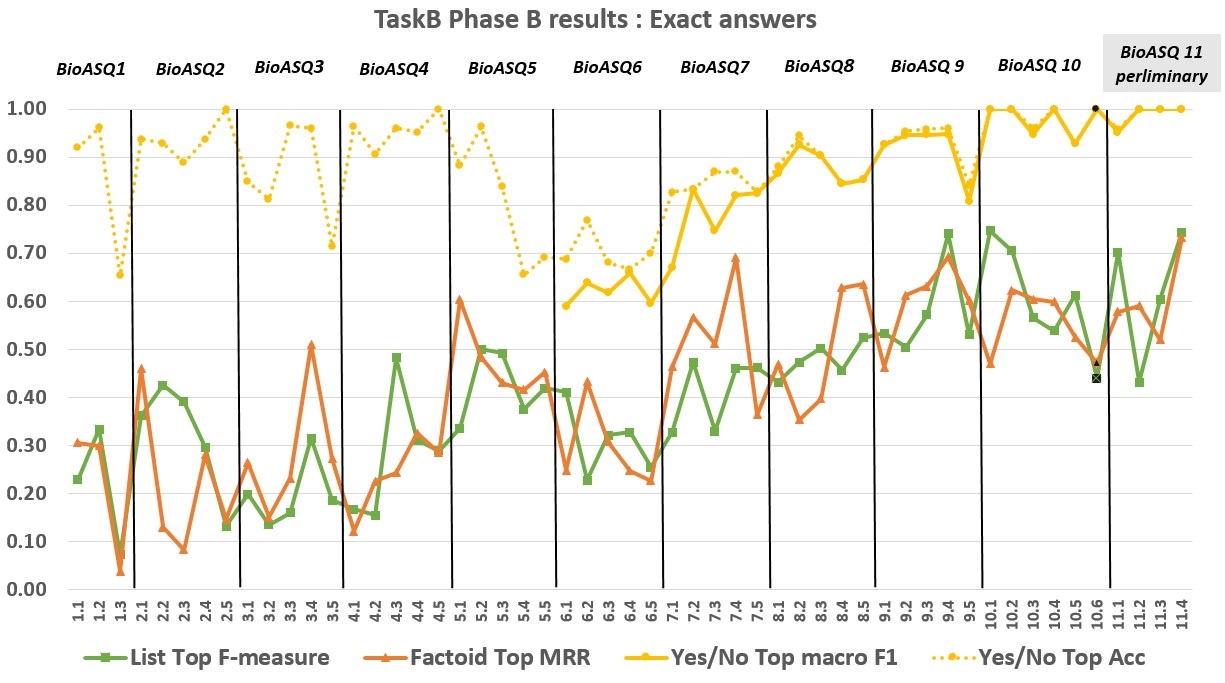}}
\caption{
The evaluation scores of the best-performing systems in task B, 
Phase B, for \textit{exact answers}, across the eleven years of the BioASQ
challenge. Since BioASQ6 the official measure for Yes/No questions is the
macro-averaged F1 score (macro F1), but accuracy (Acc) is also presented as the former official measure. The black dots in 10.6 highlight that these scores are for an additional batch with questions from new experts~\cite{nentidis2021overview}.
}\label{fig:Exact}
\end{figure*}

The top performance of the participating systems in \textit{exact answer} generation for each type of question during the eleven years of BioASQ is presented in Figure {\ref{fig:Exact}}.
The preliminary results for task 11b, reveal that the participating systems keep improving in answering all types of questions.
In batch 2, for instance, presented in Table \ref{tab:bB_res}, several systems manage to correctly answer literally all yes/no questions. This is also the case for batch 3 and batch 4.
Some improvements are also observed in the preliminary results for factoid questions compared to the previous years, but there is still more room for improvement, as done for list questions where the preliminary performance is comparable to the one of the previous year.

\subsection{Task Synergy 11}

In task Synergy 11 the participating systems were expected to retrieve documents and snippets, as in phase A of task 11b, and, at the same time, provide answers for some of these questions, as in phase B of task 11b. 
In contrast to task 11b, however, due to the developing nature of the relevant knowledge, no answer is currently available for some of the open questions. Therefore only the questions indicated to have enough relevant material gathered from previous rounds (``Answer ready'') require the submission of \textit{exact} and \textit{ideal answers} by the participating systems.

In addition, no golden documents and snippets were provided by the experts for new questions. For questions from previous rounds, on the other hand, a separate file with feedback from the experts was provided, that is elements of the documents and snippets previously submitted by the participants with manual annotations of their relevance.
Therefore, these documents and snippets, that have already been assessed and included in the feedback, were not considered valid for submission by the participants in the subsequent rounds, and even if accidentally submitted, they were not considered for the evaluation of that round. As in phase A of task 11b, the evaluation measures for document and snippet retrieval are MAP and F-measure respectively.

Regarding the \textit{ideal answers}, the systems were ranked according to manual scores assigned to them by the BioASQ experts during the assessment of systems responses as in phase B of task B~\cite{balikas13}. In this task, however, the assessment took place during the course of the task, so that the systems can have the feedback of the experts available, prior to submitting their new responses. 
For the \textit{exact answers}, which were required for all questions except the summary ones, the measure considered for ranking the participating systems depends on the question type. 
For the yes/no questions, the systems were ranked according to the macro-averaged F1-measure on the prediction of no and yes answers. 
For factoid questions, the ranking was based on mean reciprocal rank (MRR), and for list questions on mean F1-measure.

\begin{table}
    \centering
     \caption{Results for document retrieval of the first round of the Synergy 11 task.}
    \begin{tabular}{M{0.3\linewidth}M{0.14\linewidth}M{0.12\linewidth}M{0.14\linewidth}M{0.12\linewidth}M{0.12\linewidth}}
    \hline
        \textbf{System} & \textbf{Mean precision} & \textbf{Mean Recall} & \textbf{Mean F-Measure} & \textbf{MAP} & \textbf{GMAP} \\ \hline
        dmiip2             & \textbf{0.3026} & \textbf{0.3772} & \textbf{0.2803} & \textbf{0.2791} & \textbf{0.0572} \\
        dmiip4             & 0.3000 & 0.3714 & 0.2760 & 0.2788 & 0.0512 \\
        dmiip5             & 0.2667 & 0.3230 & 0.2466 & 0.2525 & 0.0578 \\
        dmiip1             & 0.2256 & 0.2668 & 0.2034 & 0.2001 & 0.0151 \\
        dmiip3             & 0.1974 & 0.2116 & 0.1741 & 0.1708 & 0.0080 \\
        bio-answerfinder   & 0.1575 & 0.1390 & 0.1244 & 0.1468 & 0.0021 \\
        bio-answerfinder-2 & 0.1575 & 0.1247 & 0.1232 & 0.1236 & 0.0011 \\
        ELECTROLBERT-3     & 0.0893 & 0.0180 & 0.0285 & 0.0212 & 0.0000 \\
        \hline 
    \end{tabular}
   \label{tab:synergy1-res}
\end{table}

Some indicative results for the Synergy task are presented in Table~\ref{tab:synergy1-res}.
The full results of Synergy 11 task are available online\footnote{\footnotesize \url{http://participants-area.bioasq.org/results/synergy\_v2023/}}. 
Overall, the collaboration between participating biomedical experts and question-answering systems allowed the progressive identification of relevant material and extraction of \textit{exact} and \textit{ideal answers} for several open questions for developing problems, such as COVID-19, Colorectal Cancer, Duchenne Muscular Dystrophy, Alzheimer's Disease, and Parkinson's Disease.   
In particular, after the completion of the the four rounds of the Synergy 11 task, enough relevant material was identified for providing an answer to about 79\% of the questions. In addition, about 42\% of the questions had at least one \textit{ideal answer}, submitted by the systems, which was considered satisfactory (ground truth) by the expert that posed the question.

\subsection{Task MedProcNER}

All in all, the top scores for each sub-task were:

\begin{itemize}
  \item \textbf{Clinical Procedure Recognition}. The BIT.UA team attained all top 5 positions with their transformer-based solution that also uses masked CRF and data augmentation. They achieved the highest F1-score, 0.7985, highest precision (0.8095) and highest recall (0.7984) . Teams Vicomtech and SINAI also obtained F1-scores over 0.75.
  \item \textbf{Clinical Procedure Normalization}. The highest F1-score (0.5707), precision (0.5902) and recall (0.5580) were obtained by Vicomtech. Teams SINAI and Fusion were also above 0.5 F1-score using token similarity techniques and a cross-lingual SapBERT, respectively. 
  \item \textbf{Clinical Procedure-based Document Indexing}. The Vicomtech team also obtained the highest F1-score (0.6242), precision (0.6371) and recall (0.6295), with the KFU NLP Team coming in second place (0.4927 F1-score). In this sub-task, all participating teams reused their systems and/or output from previous sub-tasks.
\end{itemize}

The complete results for the entity recognition, linking and document indexing are shown in tables \ref{tab:medproc-ner-results}, \ref{tab:medproc-el-results} and \ref{tab:medproc-index-results}, respectively.

\begin{table*}[!ht]
    \centering
            \caption{Results of MedProcNER Entity Recognition sub-task. The best result is bolded, and the second-best is underlined.}
    \begin{tabular}{ccccc}
    \toprule
        Team Name & Run name & P & R & F1 \\ \midrule
        BIT.UA & run4-everything & \textbf{0.8095} & \underline{0.7878} & \textbf{0.7985} \\
        BIT.UA & run0-lc-dense-5-wVal & \underline{0.8015} & \underline{0.7878} & \underline{0.7946} \\
        BIT.UA & run1-lc-dense-5-full & 0.7954 & \textbf{0.7894} & 0.7924 \\
        BIT.UA & run3-PlanTL-dense... & 0.7978 & 0.787 & 0.7923 \\
        BIT.UA & run2-lc-bilstm-all-wVal & 0.7941 & 0.7823 & 0.7881 \\
        Vicomtech & run1-xlm\_roberta... & 0.8054 & 0.7535 & 0.7786 \\ 
        Vicomtech & run2-roberta\_bio... & 0.7679 & 0.7629 & 0.7653 \\ 
        SINAI & run1-fine-tuned-roberta & 0.7631 & 0.7505 & 0.7568 \\ 
        Vicomtech & run3-longformer\_base... & 0.7478 & 0.7588 & 0.7533 \\ 
        SINAI & run4-fulltext-LSTM & 0.7538 & 0.7353 & 0.7444 \\ 
        SINAI & run2-lstmcrf-512 & 0.7786 & 0.7043 & 0.7396 \\
        SINAI & run5-lstm-BIO & 0.7705 & 0.7049 & 0.7362 \\
        KFU NLP Team & predicted\_task1 & 0.7192 & 0.7403 & 0.7296 \\
        SINAI & run3-fulltext-GRU & 0.7396 & 0.711 & 0.725 \\
        Fusion & run4-Spanish-RoBERTa & 0.7165 & 0.7143 & 0.7154 \\ 
        Fusion & run3-XLM-RoBERTA-Clinical & 0.7047 & 0.6916 & 0.6981 \\ 
        NLP-CIC-WFU & Hard4BIO...postprocessing & 0.7188 & 0.654 & 0.6849 \\ 
        NLP-CIC-WFU & Hard4BIO\_RoBERTa & 0.7132 & 0.6507 & 0.6805 \\
        Fusion & run1-BioMBERT... & 0.6948 & 0.6599 & 0.6769 \\
        Fusion & run2-BioMBERT... & 0.6894 & 0.6599 & 0.6743 \\
        Fusion & run5-Adapted-ALBERT & 0.6928 & 0.6264 & 0.658 \\
        NLP-CIC-WFU & Lazy4BIO...postprocessing & 0.6301 & 0.6002 & 0.6148 \\
        Onto-NLP & run1-...voting-filtered & 0.7425 & 0.4374 & 0.5505 \\
        Onto-NLP & run1-...voting & 0.7397 & 0.4374 & 0.5497 \\
        University Regensburg & run2-gpt-4 & 0.6355 & 0.3874 & 0.4814 \\
        saheelmayekar & predicted\_data & 0.3975 & 0.535 & 0.4561 \\
        Onto-NLP & run1-...exact\_match & 0.3296 & 0.6104 & 0.428 \\
        University Regensburg & run1-gpt3.5-turbo & 0.523 & 0.2106 & 0.3002 \\ 
    \bottomrule
    \end{tabular}

    \label{tab:medproc-ner-results}
\end{table*}

\begin{table}[!ht]
    \centering
    \caption{Results of MedProcNER Entity Linking sub-task. The best result is bolded, and the second-best is underlined.}
    \begin{tabular}{ccccc}
    \toprule
        Team Name & Run name & P & R & F1 \\ \midrule
        Vicomtech & run1-xlm\_roberta... & \textbf{0.5902} & 0.5525 & \textbf{0.5707} \\ 
        Vicomtech & run2-roberta\_bio... & \underline{0.5665} & \underline{0.5627} & \underline{0.5646} \\ 
        Vicomtech & run3-roberta\_bio... & 0.5662 & 0.5625 & 0.5643 \\ 
        Vicomtech & run5-longformer\_base... & 0.5498 & \textbf{0.558} & 0.5539 \\ 
        Fusion & run4-Spanish-RoBERTa... & 0.5377 & 0.5362 & 0.5369 \\ 
        Fusion & run1-BioMBERT... & 0.5432 & 0.516 & 0.5293 \\ 
        Fusion & run3-XLM-RoBERTA... & 0.5332 & 0.5235 & 0.5283 \\ 
        SINAI & run1-fine-tuned-roberta & 0.531 & 0.5224 & 0.5267 \\ 
        Vicomtech & run4-roberta\_bio... & 0.5248 & 0.5213 & 0.523 \\ 
        Fusion & run2-BioMBERT... & 0.5332 & 0.5105 & 0.5216 \\ 
        Fusion & run5-Adapted-ALBERT... & 0.5461 & 0.4939 & 0.5187 \\ 
        SINAI & run2-lstmcrf-512 & 0.5455 & 0.4936 & 0.5183 \\ 
        SINAI & run5-lstm-BIO & 0.5352 & 0.4898 & 0.5115 \\ 
        SINAI & run4-fulltext-LSTM & 0.5173 & 0.5047 & 0.5109 \\ 
        SINAI & run3-fulltext-GRU & 0.5079 & 0.4884 & 0.498 \\ 
        KFU NLP Team & predicted\_task2 & 0.3917 & 0.4033 & 0.3974 \\ 
        Onto-NLP & run1-pharmaconer-top1 & 0.2742 & 0.508 & 0.3562 \\ 
        Onto-NLP & run1-pharmaconer-voter & 0.2723 & 0.5044 & 0.3536 \\ 
        Onto-NLP & run1-cantemist-top1 & 0.2642 & 0.4895 & 0.3432 \\ 
        Onto-NLP & run1-ehr-top1 & 0.263 & 0.4873 & 0.3416 \\ 
        BIT.UA & run4-everything & 0.3211 & 0.3126 & 0.3168 \\ 
        BIT.UA & run3-PlanTL-dense... & 0.3188 & 0.3145 & 0.3166 \\ 
        BIT.UA & run0-lc-dense-5-wVal & 0.318 & 0.3126 & 0.3153 \\ 
        BIT.UA & run1-lc-dense-5-full & 0.3143 & 0.3121 & 0.3132 \\ 
        BIT.UA & run2-lc-bilstm-all-wVal & 0.3133 & 0.3087 & 0.311 \\ 
        University Regensburg & run2-gpt-4 & 0.4304 & 0.1282 & 0.1976 \\ 
        University Regensburg & run1-gpt-3.5-turbo & 0.4051 & 0.0749 & 0.1264 \\ 
        \bottomrule
    \end{tabular}
    \label{tab:medproc-el-results}
\end{table}

\begin{table}[!ht]
    \centering
        \caption{Results of MedProcNER Indexing sub-task. The best result is bolded, and the second-best is underlined.}
    \begin{tabular}{ccccc}
    \toprule
        Team Name  & Run name & P & R & F1 \\ \midrule
        Vicomtech & run5\_roberta\_bio... & \underline{0.619} & \textbf{0.6295} & \textbf{0.6242} \\ 
        Vicomtech & run4\_xlm\_roberta... & \textbf{0.6371} & 0.6109 & \underline{0.6239} \\ 
        Vicomtech & run1\_roberta\_bio... & 0.6182 & \textbf{0.6295} & 0.6238 \\ 
        Vicomtech & run3\_longformer... & 0.6039 & \underline{0.6288} & 0.6161 \\ 
        Vicomtech & run2\_roberta\_bio... & 0.5885 & 0.5917 & 0.5901 \\ 
        KFU NLP Team & predicted\_task3 & 0.4805 & 0.5054 & 0.4927 \\ 
        BIT.UA & run3-PlanTL-dense... & 0.3544 & 0.3654 & 0.3598 \\ 
        BIT.UA & run4-everything & 0.3551 & 0.3619 & 0.3585 \\ 
        BIT.UA & run0-lc-dense-5-wVal & 0.3517 & 0.3619 & 0.3567 \\ 
        BIT.UA & run1-lc-dense-5-full & 0.3475 & 0.3612 & 0.3542 \\ 
        BIT.UA & run2-lc-bilstm-all-wVal & 0.3484 & 0.3593 & 0.3537 \\ 
        University Regensburg & run2-gpt-4 & 0.5266 & 0.1811 & 0.2695 \\ 
        University Regensburg & run1-gpt3.5-turbo & 0.506 & 0.1083 & 0.1785 \\ 
        \bottomrule
    \end{tabular}
    \label{tab:medproc-index-results}
\end{table}

Overall, the performance of the systems presented for the MedProcNER shared task is very diverse, with scores ranging from 0.759 F-score (by the BIT.UA team on the entity recognition task) to 0.126 (University of Regenburg on the entity linking task). This gap evidences mainly two things: the multitude of approaches and the difficulty of the corpus. 
On the one hand, the systems presented for the task were very varied. Even amongst BERT-based models, participants tried different strategies such as using models pre-trained on different domains (biomedical, clinical) and languages (Spanish, multilingual), implementing different pre/post-processing techniques, data augmentation and using multiple output layers (CRF, GRU, LSTM). Again, it is remarkable that one of the participants (Samy Ateia from the University of Regensburg) used GPT3.5 (ChatGPT) and GPT4 for their submissions. Even though the overall performance is not too good (especially in terms of recall), this is partly to be expected since the system was fine-tuned for the task using a few-shot approach.
On the other hand, the Gold Standard corpus is very varied in terms of mentions, with many mentions being quite long and descriptive (especially surgical mentions). Additionally, the text documents span multiple medical specialties, which introduces not only more variety in clinical procedures but also possible ambiguities due to the use of specialized abbreviations. In the future, we will expand the corpus with more annotated documents to address this issue.

Compared to last year's DisTEMIST task, which had a very similar setting, results are overall a bit higher but still quite similar. In terms of named entity recognition methodologies, transformers and BERT-like models were the most popular in both tasks, with RoBERTa not only being the most widely used but also achieving some of the best results. 
In the entity linking sub-task systems that use SapBERT seem to have gained popularity, being used by at least 3 teams, including the top-scoring system, with very good results. In contrast, in last year's DisTEMIST only one team (HPI-DHC) used it, and actually achieved the best entity linking score using an ensemble of SapBERT and TF-IDF with re-ranking and a training data lookup.

\section{Conclusions}
\label{sec:conclusion}

This paper provides an overview of the eleventh BioASQ challenge.
This year, the challenge consisted of three tasks: (1) Task 11b on biomedical semantic question answering in English and (2) task Synergy 11 on question answering for developing problems, both already established from previous years of the challenge, and (3) the new task MedProcNER on retrieving medical procedure information from medical content in Spanish.

The preliminary results for task 11b reveal some improvements in the performance of the top participating systems, mainly for yes/no and factoid answer generation. However, room for improvement is still available, particularly for factoid and list questions, where the performance is less consistent.

The new edition of the Synergy task in an effort to enable a dialogue between the participating systems with biomedical experts revealed that state-of-the-art systems, despite still having room for improvement, can be a useful tool for biomedical experts that need specialized information for addressing open questions in the context of several developing problems.

The new task MedProcNER introduced three new challenging subtasks on annotating clinical case reports in Spanish. Namely, Named Entity Recognition, Entity Linking, and Semantic Indexing for medical procedures. Due to the importance of semantic interoperability across data sources, SNOMED CT was the target terminology employed in this task, and multilingual annotated resources have been released. This novel task on medical procedure information indexing in Spanish highlighted the importance of generating resources to develop and evaluate systems that (1) effectively work in multilingual and non-English scenarios and (2) combine heterogeneous data sources.

The ever-increasing focus of participating systems on deep neural approaches, already apparent in previous editions of the challenge, is also observed this year.
Most of the proposed approaches built on state-of-the-art neural architectures (BERT, PubMedBERT, BioBERT, BART etc.) adapted to the biomedical domain and specifically to the tasks of BioASQ. 
This year, in particular, several teams investigated approaches based on Generative Pre-trained Transformer (GPT) models for the BioASQ tasks. 

Overall, several systems managed competitive performance on the challenging tasks offered in BioASQ, as in previous versions of the challenge, and the top performing of them were able to improve over the state-of-the-art performance from previous years.
BioASQ keeps pushing the research frontier in biomedical semantic indexing and question answering for eleven years now, offering both well-established and new tasks. 
Lately, it has been extended beyond the English language and biomedical literature, with the tasks MESINESP \cite{luis2020overview}, DisTEMIST \cite{amiranda2022overview}, and this year with MedProcNER. 
In addition, BioASQ reaches a more and more broad community of biomedical experts that may benefit from the advancements in the field. This has been done initially for COVID-19, through the introductory versions of Synergy, and was later extended into more topics with the collaborative batch of task 10b and the extended version of Synergy 11, introduced this year.
The future plans for the challenge include a further extension of the benchmark data for question answering through a community-driven process, extending the community of biomedical experts involved in the Synergy task, as well as extending the resources considered in the BioASQ tasks, both in terms of documents types and language.

\section{Acknowledgments}
Google was a proud sponsor of the BioASQ Challenge in 2022. 
The eleventh edition of BioASQ is also sponsored by Ovid.
Atypon Systems Inc. is also sponsoring this edition of BioASQ. 
The MEDLINE/PubMed data resources considered in this work were accessed courtesy of the U.S. National Library of Medicine.
BioASQ is grateful to the CMU team for providing the \textit{exact answer} baselines for task 11b, as well as to Georgios Moschovis and Ion Androutsopoulos, from the Athens University of Economics and Business, for providing the \textit{ideal answer} baselines.
The MedProcNER track was partially funded by the Encargo of Plan TL (SEDIA)  to  the Barcelona Supercomputing Center.  Due to  the  relevance  of  medical procedures for implants/devices specially in the case cardiac diseases this  project  is also supported by the  European  Union’s  Horizon  Europe  Coordination  \&  Support  Action  under  Grant Agreement No 101058779 (BIOMATDB) and DataTools4Heart Grant Agreement No. 101057849. We also acknowledge the  support  from  the AI4PROFHEALTH project (PID2020-119266RA-I00).
%
%
%
\bibliographystyle{splncs04}
\bibliography{BioASQ11.bib}

\end{document}